\documentclass[runningheads]{llncs}
\usepackage[dvipdfmx]{graphicx}
\usepackage{amsmath,amssymb}
\usepackage{arydshln}
\usepackage{color}

\newcommand{\bp}{{\bf p}}

\begin{document}
\title{Proportion Estimation by Masked Learning from Label Proportion}
\titlerunning{Proportion Estimation by Masked Learning from Label Proportion}


\author{
Takumi Okuo\inst{1} \and
Kazuya Nishimura\inst{1} \and
Hiroaki Ito\inst{2} \and
Kazuhiro Terada\inst{2} \and
Akihiko Yoshizawa\inst{2} \and
Ryoma Bise\inst{1}
}
\institute{
Kyushu University, Fukuoka, Japan \email{takumi.okuo@humna.ait.kyushu-u.ac.jp} 
 \and Kyoto University Hospital, Kyoto, Japan
}
\authorrunning{T. Okuo et al.}

% index{Okuo, Takumi} 
% index{Nishimura, Kazuya} 
% index{Ito, Hiroaki} 
% index{Terada, Kazuhiro} 
% index{Yoshizawa, Akihiko} 
% index{Bise, Ryoma} 

\maketitle              % typeset the header of the contribution
\begin{abstract}
The PD-L1 rate, the number of PD-L1 positive tumor cells over the total number of all tumor cells, is an important metric for immunotherapy. 
This metric is recorded as diagnostic information with pathological images.
In this paper, we propose a proportion estimation method with a small amount of cell-level annotation and proportion annotation, which can be easily collected.
Since the PD-L1 rate is calculated from only `tumor cells' and not using `non-tumor cells', we first detect tumor cells with a detection model. 
Then, we estimate the PD-L1 proportion by introducing a masking technique to `learning from label proportion.' In addition, we propose a weighted focal proportion loss to address data imbalance problems.
Experiments using clinical data demonstrate the effectiveness of our method. Our method achieved the best performance in comparisons.

\end{abstract}
\section{Introduction}
\begin{figure}[t]
    \centering
    \includegraphics[width=\linewidth]{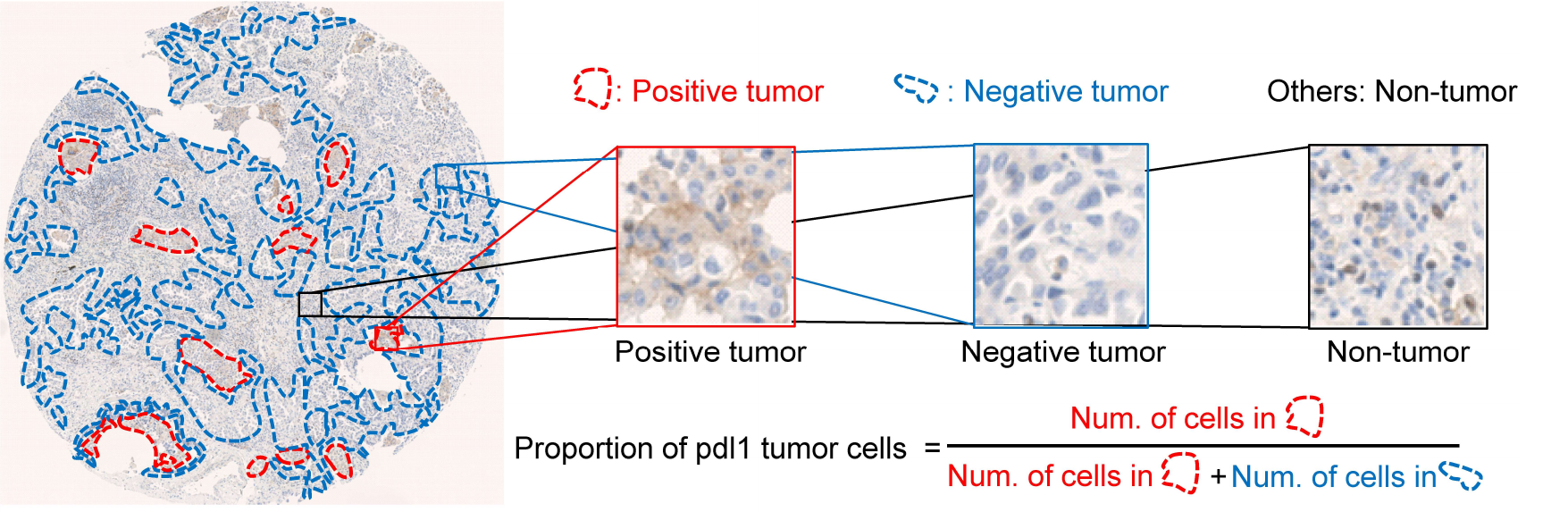}
    \caption{Illustration of a core image in PD-L1 test. Blue dots regions indicate negative tumors, and red indicates positive tumors. Ideally, the proportion of positive tumor cells is calculated by counting and classifying all cells; however, pathologists roughly estimate the ratio subjective without counting or segmenting.
    }
    \label{fig:intro}
\end{figure}

The proportional information of cancer subtypes is recorded as diagnostic information with pathological images in many diagnoses, such as programmed cell death ligand-1 (PD-L1) diagnosis~\cite{Liu2021,Widmaier2020}, chemotherapy, and lung cancer diagnosis~\cite{tokunaga2020negative,yoshizawa2011impact}. 
For example, the PD-L1 test is conducted to check if cancer immunotherapy will be helpful for a patient.
Fig. \ref{fig:intro} shows an example image (called core image) captured by a whole slide scanner.
In the core image, over 10 thousand cells belong to three classes; positive cancer, negative cancer, and non-tumor cells. 
The PD-L1 rate is calculated by the number of positive tumor cells over all of the tumor cells in a tissue. Note that it does not include non-tumor cells.
Counting all cells in the core image is almost impossible; thus, pathologists roughly estimate the PD-L1 rate without counting in clinical.
Therefore, there is a demand to develop an automatic proportion estimation method.

A simple solution is to detect positive and negative tumor cells by a deep detection model \cite{redmon2016you} and calculate the proportion from the detection results.
However, this approach does not use the useful information, PD-L1 rate, which has already been recorded as diagnosis information and requires a certain amount of cell-level annotation for positive and negative tumor cells.
A small amount of annotation can be collected, but the performance of a network trained using insufficient training data is worse.
Large datasets with sufficient variability must be collected to address inter-tumor heterogeneity and different staining characteristics, but it is time-consuming and labor-intensive.

Another solution is to estimate the proportion by regression or classification directly.
The approach can train a network using proportion information without additional annotation since PD-L1 rates are recorded in clinical settings.
However, these methods have two drawbacks. A PD-L1 rate is a `partial' proportion of all cells, i.e., it is calculated only from tumor cells but ignores non-tumor cells. This ambiguity caused by without using the non-tumor region information makes training difficult to estimate proportion.
Another disadvantage is poor interpretability for estimation.
When pathologists use the estimated results in clinical, they check the cell distributions of three classes to know how the rate is calculated.
A class activation map (CAM)-based approach~\cite{zhou2016learning}, which visualizes contributed pixels to the output, is unsuitable for this task because class distributions of all tumor regions are required to estimate the proportion.

This paper proposes a proportion estimation method to estimate the proportion of PD-L1 tumor cells and output the distributions of three classes in a core image while keeping annotation costs as low as possible.
As discussed above, estimating the proportion without tumor region information is difficult. Thus, we made a small amount of training data to train a cell detection network that detects tumor or non-tumor cells.
This enables us to make a tumor region mask, which is used for estimating the proportion (PD-L1 rate).
We proposed a masked `Learning from Label Proportions (LLP)' that estimates the proportion by using the tumor cell region mask predicted by the cell detection network.
In addition, we propose a weighted focal proportion loss to address data imbalance problems, where data imbalance often occurs in medical image applications.
Experiments using clinical data demonstrate the effectiveness of our method. Our method achieved the best performance in comparison.

\section{PD-L1 tumor proportion estimation}
\label{sec:method}
Our method aims to train a network that estimates the proportion $r$ of PD-L1 tumor cells in core images $I$. 
We use three types of annotation for the training. 
One is a cell position label, the second is a tumor cell region, and the third is the proportion label.
For proportion labels, a proportion is often recorded as interval sections in clinical since pathologists roughly estimate the PD-L1 rate subjective without counting.
For example, PD-L1 rates are either `0 to 0.01', `0.01 to 0.25', `0.25 to 0.5', `0.5 to 0.75', or `0.75 to 1.00'. 

\begin{figure}[t]
    \centering
    \includegraphics[width=\linewidth]{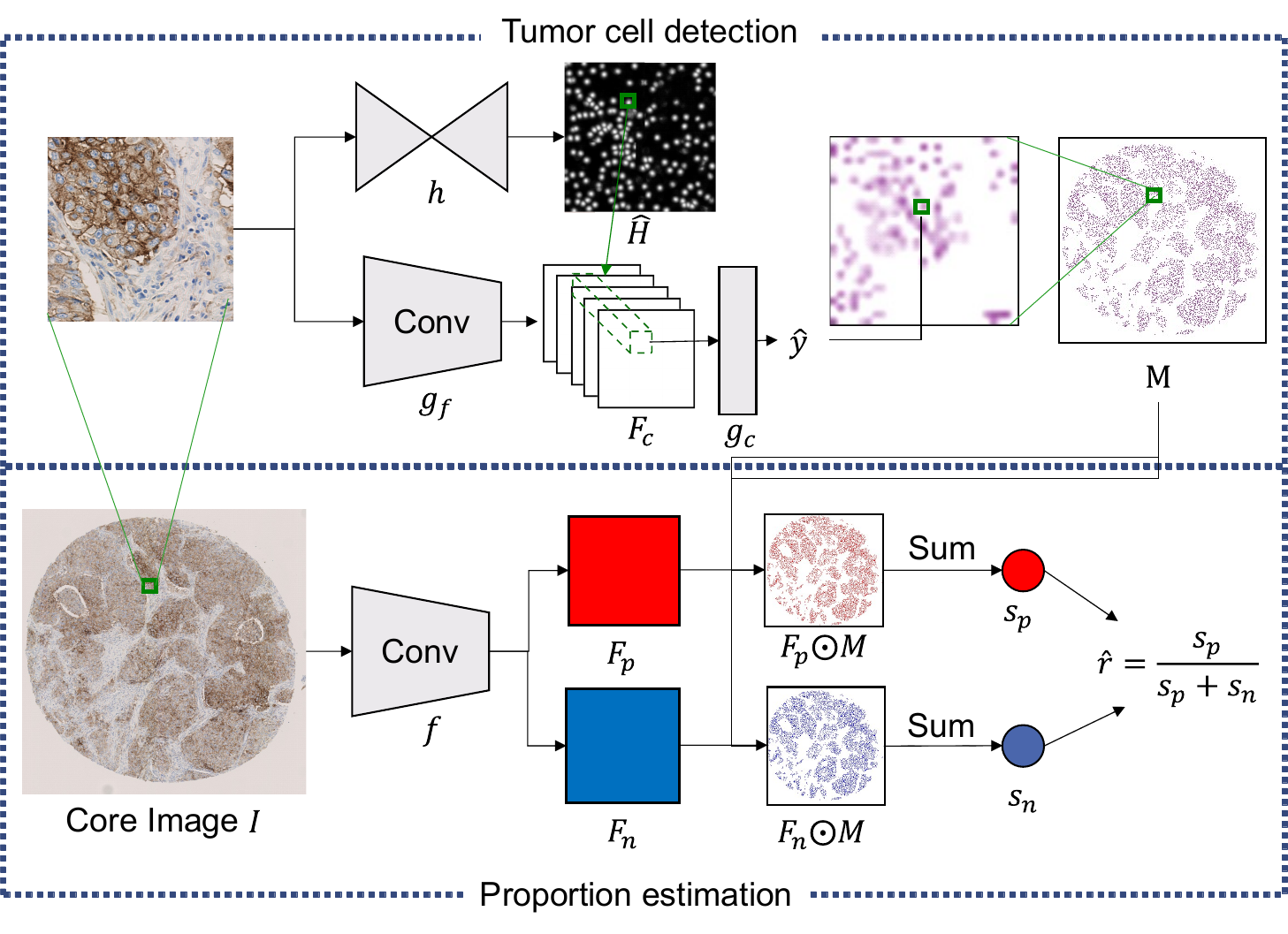}
    \caption{Overview of proposed method. Top: cell detection network, and Bottom: proportion estimation network.}
    \label{fig:overview}
\end{figure}

Fig. \ref{fig:overview} shows an overview of the proposed method. 
Our method was designed as a two-stage structure to effectively use a small amount of cell-level annotation and a large amount of tissue-level annotation (proportion).
We first detect tumor cells in core image $I$ to produce tumor mask $M$.
Then, we estimate the proportion of PD-L1-positive cells of an input image $I$ using mask $M$. 

%\vskip%\baselineskip
\noindent
{\bf Tumor cell detection:}
Fig. \ref{fig:overview} (Top) shows the overview of the tumor cell detection network, which consists of a cell detection model $h$ and a classification model $g$.
For the cell detection model $h$, we follow the heatmap-based cell detection method \cite{nishimura2021weakly}, which produces a heatmap of all cells, where the coordinates of the peaks in the maps indicate the centroid positions of cells.
For the classification model $g$, we propose a two-stage cell detection method inspired by bounding box-based general object detectors~\cite{ren2015faster}.
Since cell shapes are similar between tumor and non-tumor cells, it is difficult to accurately identify their class from local information.
To use the information of surrounding cells effectively, we first extract global feature by feature extractor $g_f$ and then classify cells based on the extracted feature, which contains the surrounding information.

We first train the cell detection network $h$ using the training data of cell positions, where $h$ produces the cell position heatmap for input image $I$~\cite{nishimura2021weakly}.
To train $h$, the ground truth of the heatmap $H$ is generated using the given training data of cell positions. The network $h$ is trained by minimizing the MSE Loss $L_d=\|H-\hat{H}\|^2$, where $\hat{H}=h(I)$ is an estimated heatmap by $h$. The peak points in $\hat{H}$ are detected as the cell positions in a testing phase, denoted as $\{\hat{\bp}_i\}_{i=1}^{N_d}$, where $N_d$ is the number of detected cells in $I$. Note that $\hat{H}$ only contains the cell position information but not the tumor class information.

For each detected cell position $\hat{\bp}_i$, the tumor classification network $g$ (consists of $g_f$ and $g_c$) estimates its class $\hat{y}_i$. The feature extractor $g_f$ extracts a feature map $F_t$ from $I$, and then the fully connected (Fc) layer $g_c$ estimates the tumor class $\hat{y}_i \in [0,1]$ for each cell position by inputting the feature vector $F_t(\hat{\bp}_i)$ at pixel $\hat{\bp}_i$, where $\hat{y}_i>0.5$ indicates a tumor cell, otherwise, a non-tumor cell.
To train this network $g$, we use the binary cross-entropy loss between the predicted score $\hat{y}_i$ and the ground truth $y_i$, where the loss is calculated at only the detected cell positions $\{\hat{\bp}_i\}_{i=1}^{N_d}$. The detection results are denoted as $\hat{\mathcal{P}} = \{\hat{\bp_i}, \hat{y}_i\}_{i=1}^{N_d}$.

We generate a tumor cell mask $M$ using detection results $\hat{\mathcal{P}}$. In $M$, pixels around the tumor cell positions $ \hat{\mathcal{P}}_c=\{ \hat{p}_i | \hat{y}_i > 0.5\}$ takes 1, otherwise 0, where the distance from the detected position to a positive pixel is less than $\alpha$.
This mask is used to estimate the proportions.

\begin{figure}[t]
    \centering
    \includegraphics[width=\linewidth]{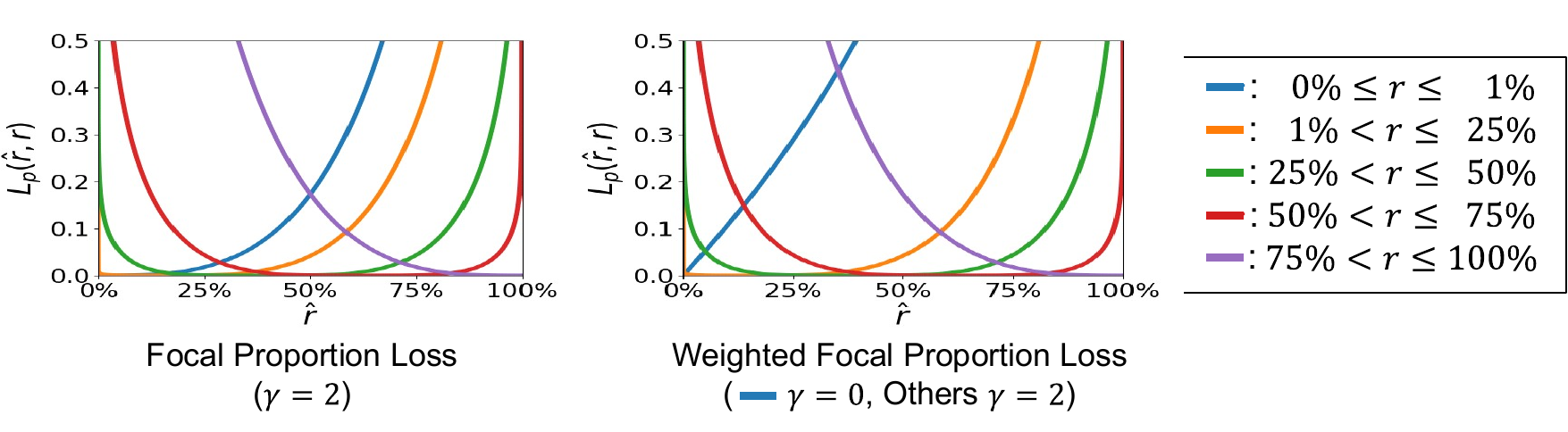}
    \caption{
    Left: focal proportion loss. Right: weighted focal proportion loss.
    }
    \label{fig:loss}
\end{figure}

%\vskip%\baselineskip
\noindent
{\bf PD-L1 proportion estimation:}
The proportion estimation network $f$ estimates the proportion of PD-L1 positive cells among tumor cells.
As shown in Fig. \ref{fig:loss}, the feature extractor of $f$ extracts the positive map $F_p$ and negative map $F_n$ by inputting $I$. These feature maps are masked to calculate the pixels on only the tumor regions, denoted as $F_p \odot M$ and $F_n \odot M$, respectively, which indicate the maps of the positive/negative `tumor' cell position maps.
Then, the estimated proportion $\hat{r}$ is calculated from these masked maps. The positive score $s_p$ and negative score $s_n$, which indicate the numbers of positive or negative tumor cells in the image, are defined as the sum of the pixel values in $F_p \odot M$ and $F_n \odot M$.
The PD-L1 tumor proportion is calculated by $\hat{r} = \frac{s_p}{s_p + s_n}$. The network is trained using the estimated proportion $\hat{r}$ and the ground truth $r$. The details of the loss function are proposed below.

As discussed in the introduction, the PD-L1 rate is given as a proportion interval; pathologists give either `0 to 0.01', `0.01 to 0.25', `0.25 to 0.5', `0.5 to 0.75', or `0.75 to 1.00' for a core image.
In previous works, the loss for proportion interval \cite{bortsova2018deep} outperforms cross-entropy. This loss mitigates the problem of overfitting by sacrificing strictness but at the expense of discriminability.
In our problem setting, there are different intervals; the interval in `0 to 0.01' is much smaller than that of others, and there is data imbalance, e.g., the number of core images belonging to `0.5 to 0.75' is much fewer than that of `0 to 0.01'. These issues make training difficult.

We thus propose a weighted focal proportion loss that can mitigate these issues. This loss is designed inspired by the focal loss~\cite{lin2017focal}, which has been widely used for imbalanced data classification. The focal loss is a dynamically scaled cross-entropy loss, where the scaling factor decays to zero as confidence in the correct class increases~\cite{lin2017focal}.
We introduce this idea into the proportion loss~\cite{ardehaly2017co}, widely used in LLP. 
Let us denote $r_k, (k=1,2)$ are the proportion of the positive and negative cells. The weighted focal proportion loss is defined as:
\begin{align}
    WFL = -|r-\hat{r}|^{\gamma}\left(\sum_{k} r_k\log \frac{r_k}{\hat{r_k}}\right),
\end{align}
where $r$ is the ground truth of the proportion (PD-L1 rate), which takes the mean of the interval; e.g., $r=0.375$ for `0.25 to 0.5'. $\hat{r}$ is the estimated proportion, and $\gamma$ is a hyper-parameter.
$\sum_{k} r_k\log \frac{r_k}{\hat{r_k}}$ indicates the KL-divergence between the truth and estimated proportion (proportion loss), $r_1=r$ indicates the proportion of the positive cells, and $r_2=1-r_k$ indicates that of the negative one. 
$|r-\hat{r}|^{\gamma}$ indicates that the loss decreases when the estimation becomes the correct value.

Figure~\ref{fig:loss} (Left) shows the plot of the focal proportion loss for each proportion interval when $\gamma = 2$. In this graph, the loss for `0 to 0.01' (blue line) has low gradients, and the loss around 3 \% is lower than that of the neighboring interval `0.01 to 0.25'.
It makes training difficult to distinguish the proportions from the neighbor intervals.
This is because the interval of `0 to 0.01' is much smaller than the others. Therefore, we add the weight for the hyper-parameter $\gamma$, where smaller $\gamma$ gives more significant gradients: we use $\gamma = 0$ for `0 to 0.01' and $\gamma = 2$ for other intervals.
Figure~\ref{fig:loss} (Left) shows the weighted focal proportion loss curves. The blue curve (`0 to 0.01') has larger gradients, and thus this makes training easy to identify the proportion in either `0 to 0.01' or `0.25 to 0.5'.

\section{Experiments}

\noindent
{\bf Implementation details:}
For tumor cell detection, we used U-net structure \cite{ronneberger2015u} for the cell detector $h$, and Resnet-50 \cite{he2016deep} for the feature extractor $g_f$.
The Adam optimizer \cite{kingma2014adam} is adopted with learning rates of 0.001 for $h$ and 0.0002 for $g$, and the batch sizes were 8 and 16 for $h$ and $g$, respectively. Random rotation and flipping were applied for augmentation during classification training. 

For proportion estimation, we used Resnet-18 \cite{he2016deep} pre-trained on  Imagenet.
The Adam optimizer \cite{kingma2014adam} is adopted with a learning rate of 0.0001. The batch size was 16, and the epoch was 100. We used early stopping with a patience of 30 to stop training.  
We used random rotation and horizontal and vertical flipping for augmentation.
We set the hyperparameter $\gamma$, which controls the slope of the weighted focal loss function, to 0 for 0 to 1 \% and 2 for the other sections. 

\noindent
{\bf Dataset:}
For tumor detection, we used 58 core images of patients with about 10000 $\times$ 10000.
The tumor or non-tumor regions, pathologists manually annotated, where the average of the labeled region in one core image is $5.71\%$.
For two core images, the cell position is also annotated. The total number of cells is 8000.
Note that if we were to train the network using only the supervised data of cell detection, a huge amount of annotation would be required. In contrast, we can train the network using proportion labels and only a small amount of cell-level annotation ($5.71\%$ area in a tissue). Proportion labels are available from clinical records, and no public datasets have proportion annotations.

\begin{figure}[t]
    \centering
    \includegraphics[width=\linewidth]{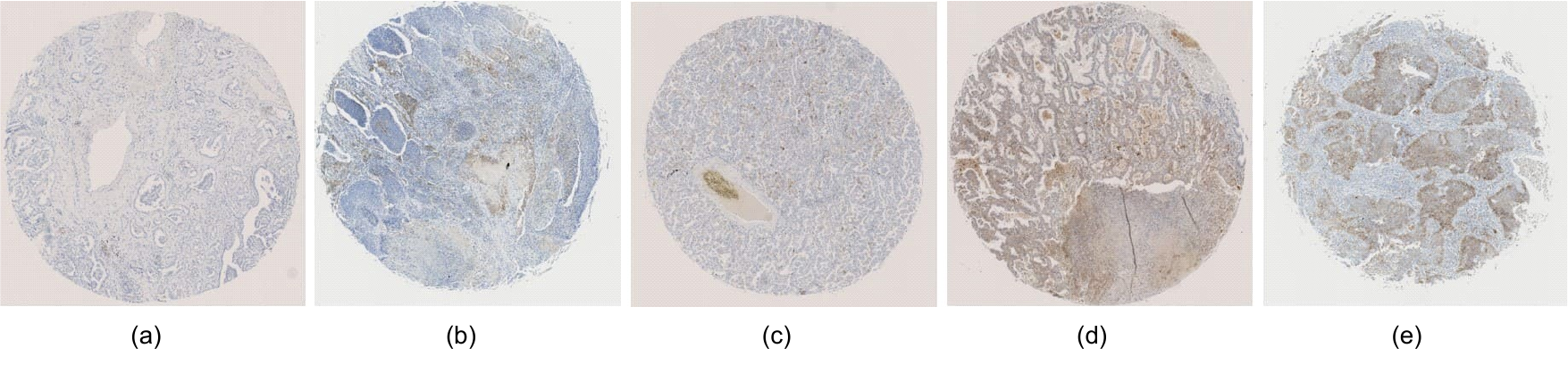}
    \caption{Example of core images in cases of (a) 0 to 1\%, (b) 1 to 25\%, (c) 25 to 50\%, (d) 50 to 75\%, (e) 75 to 100\%.}
    \label{fig:dataset}
\end{figure}

For proportion estimation, we used 606 core images, where the proportion interval is labeled for each core; either `0 to 0.01', `0.01 to 0.25', `0.25 to 0.5', `0.5 to 0.75', or `0.75 to 1.00'.
We resized the images to 2048 $\times$ 2048 to input the network.
Fig. \ref{fig:dataset} shows examples of core images.
The images gradually turn brown along with a larger proportion of PD-L1 positive cells. However, it is necessary to observe whether the brown cells are tumors and whether the membranes are dyed.
This means that it is not so easy for color-based methods.
We performed 4-fold cross-validation and evaluated the average performance metrics.

\begin{table}[t]
  \centering
  \caption{Performance of proportion estimation by comparative methods.}
    \label{tab:quan}
    \resizebox{\linewidth}{!}{%
    \begin{tabular}{c|c|cccccccc} \hline
        Method & w Mask & 0-1 \% & 1-25 \% & 25-50 \% & 50-75 \% & 75-100 \% &mRecall &mPrecision & mF1 \\ \hline \hline
        Det && 0.832 & 0.587 & 0.062 & 0.071 & 0.568 & 0.479 & 0.424 & 0.429 \\ \hdashline 
        Class \cite{he2016deep} & & 0.858 & 0.433 & 0.312 & 0.411 & 0.682 & 0.539 & 0.526 & 0.524 \\
        O-Reg \cite{cheng2008neural} && 0.811 & 0.510 & 0.500 & 0.411 & 0.568 & 0.560 & 0.541 & 0.532 \\ 
        Prop w/o mask \cite{ardehaly2017co} && 0.381 & 0.779 & 0.438 & 0.339 & 0.750 & 0.537 & 0.510 & 0.451 \\
        Ours w/o mask && 0.627 & 0.529 & 0.125 & 0.232 & 0.795 & 0.462 & 0.442 & 0.427 \\ \hdashline 
        LPI Loss \cite{bortsova2018deep} &\checkmark & 0.187 & {\bf 0.808} & 0.188 & 0.429 & 0.682 & 0.458 & 0.519 & 0.392\\ 
        Prop &\checkmark & 0.474 & 0.760 & 0.438 & 0.464 & 0.795 & 0.586 & 0.561 & 0.516\\
        WProp &\checkmark & 0.795 & 0.683 & 0.312 & 0.393 & 0.795 & 0.596 & 0.552 & 0.558\\
        WProp + List &\checkmark & 0.837 & 0.692 & 0.375 & 0.339 & 0.705 & 0.590 & 0.567 & 0.565\\
        FocalProp &\checkmark & 0.078 & 0.731 & {\bf 0.562} & {\bf 0.482} & {\bf 0.818} & 0.534 & 0.538 & 0.404\\ 
        Ours & \checkmark & {\bf 0.878} & 0.663 & 0.375 & 0.393 & 0.795 & {\bf 0.621} & {\bf 0.606} & {\bf 0.603}\\ 
        \hline
    \end{tabular}
    }
\end{table}

\noindent
{\bf Evaluation:}
To confirm the effectiveness of our proportion estimation with mask, we compared our method with ten baseline methods; 1) the detection-based method (Det) modified our cell detection network for 3 classes of cell detection: positive tumor, negative tumor, and non-tumor. The proportion of PD-L1 is calculated from the number of detected positive and negative tumor cells, where the proportion labels were not used to train this method.
2) Classification (Class)~\cite{he2016deep}, which directly classifies the core image into five classes (proportion intervals) with cross-entropy loss.
3) Ordinal regression (O-Reg)~\cite{cheng2008neural}. which estimates used the proportion using the loss function from \cite{cheng2008neural}.
4)Prop w/o mask~\cite{ardehaly2017co}, which estimates continuous values with proportion loss without using the mask.
5) Ours without mask (Ours w/o mask), which uses the weighted focal proportion loss without tumor cell detection.
The above 2)-4) methods did not use the tumor region mask $M$, i.e., the networks were trained to produce the PD-L1 rate directly from the entire image.
As an ablation study, the following five methods are introduced into our framework, which uses the mask $M$ and estimates the proportion from the masked maps.
6) LPI Loss~\cite{bortsova2018deep}.
7) Proportion loss (Prop)~\cite{ardehaly2017co}.
8) Weighted proportion loss (WProp), where the proportion loss is weighted by the length of the interval.
9) WProp + List, which introduces listnet~\cite{cao2007learning} to the weighted proportion loss.
10) Forcal proportion loss (FocalProp), which introduces the focal loss~\cite{lin2017focal} into the proportion loss. This is proposed by us.
11) Ours, which uses the weighted focal proportion loss.

\begin{figure}[t]
    \centering
    \includegraphics[width=\linewidth]{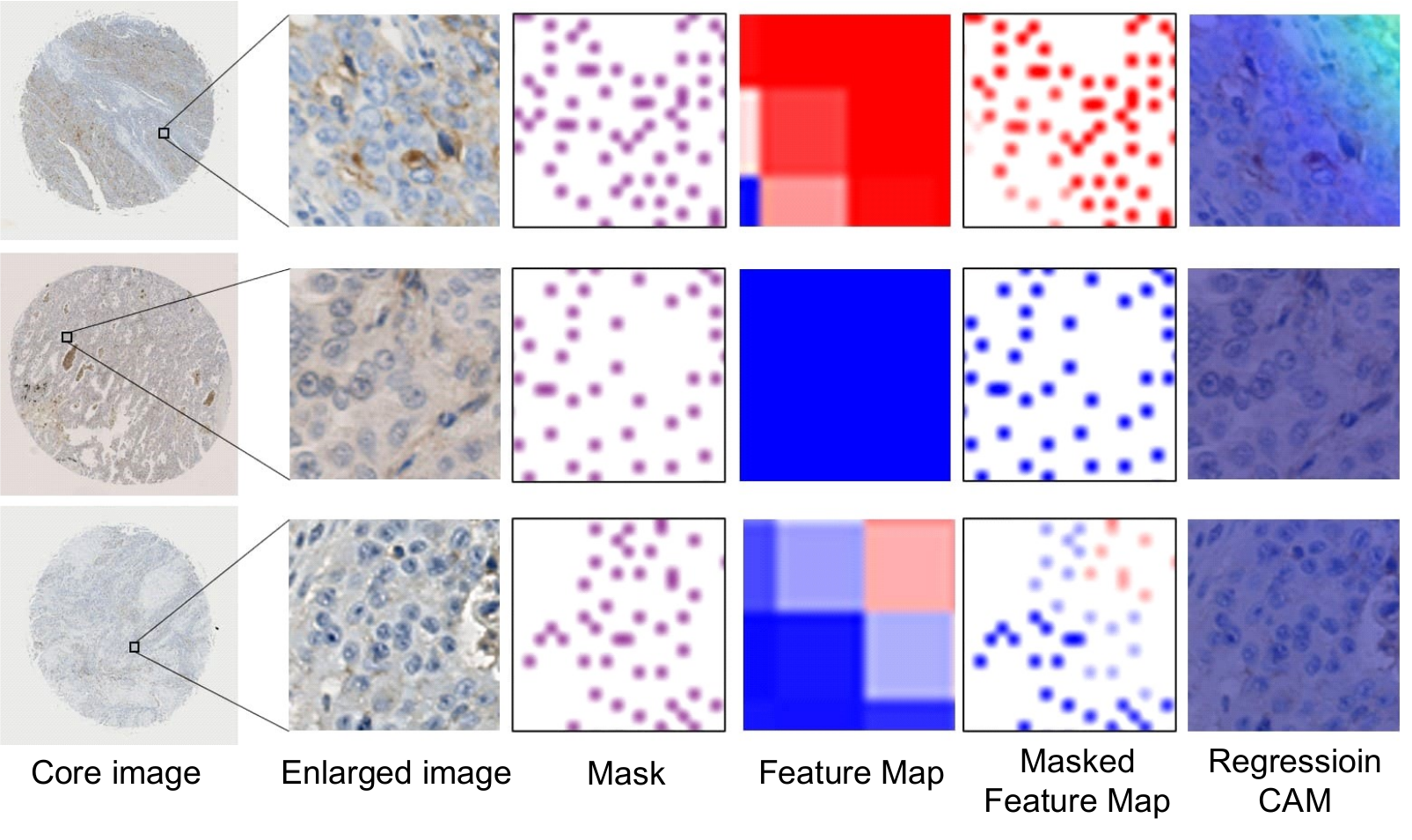}
    \caption{Visualization by intermediate outputs and CAM-based method~\cite{zhou2016learning}. Red indicates a positive, and blue indicates a negative.
    }
    \label{fig:visualize}
\end{figure}

Table \ref{tab:quan} shows the mean of precision, recall, and f1 score for each method.
Our method achieved the best performance compared to the other methods.
Positive and negative tumor cells have various appearances depending on staining properties and patient variation. 
Det did not work well for three-class classifications as the positive and negative tumor cells have similar shapes, as shown in Fig.~\ref{fig:visualize}. To achieve accurate detection, a large amount of training data is required. 
Comparing Prop w/o mask and Prop, Prop is better than Prop w/o mask. 
Since both methods use the same loss function, it shows that the tumor mask contributes to improving performance.
The accuracy of `0 to 1 \%' in FocalProp is much worse because this loss teats all interval sections the same even though the interval length is different, as discussed in Section~\ref{sec:method}.
The proposed method outperformed other mask-based methods because weighted focal loss can handle imbalances (interval and data) while maintaining discriminative ability.

Fig. \ref{fig:visualize} shows examples of estimated intermediate positive and negative feature maps, which can be used for the interpretability of the network; pathologists can understand how the AI classifies the cells to estimate the proportion.
The mask (the 3rd column) shows the cell detection results of tumor cells (including both positive and negative).
The feature maps show the estimation results of positive (red) or negative (blue) classification, in which the feature map has a low resolution than the original image. 
A masked feature map is generated by combining the mask and feature map.
Regression CAM shows the activation map from the regression network, where red indicates the pixels contributing to the network output.
The first and second rows are examples of successfully estimated cases, which pathologists confirmed.
In these cases, both of them are stained as brown. However, the class of cells is different; positives in the 1st column and negatives in the 2nd one, which is defined by the staining pattern even though they are brown. Our method successfully classified such difficult cases.
The third row is an example of a miss-classified case, which is a difficult case. Actually, all of them are positive because their membrane is slightly stained by light brown. However, our method miss-classified them as negative. This is a difficult case, even for medical doctors.
The CAM is meaningless in this task because all tumor regions are used to calculate the proportions.

\section{Conclusion}
We propose a proportion estimation method that can estimate the partial proportion (about only tumor cells) by using a tumor mask and address the imbalanced (interval and data) issues by our weighted focal proportion loss.
We first detect tumor cells and generate a tumor mask.
Then, we estimate the PD-L1 tumor proportion among tumor cells. 
By applying the mask, we could represent intermediate output for PD-L1 positive and negative, in which the visualization is useful for pathologists in clinical.
In the experiments, our method outperforms other comparisons and achieves state-of-the-art performance.

\noindent
{\bf Acknowledgements:} This work was supported by JSPS KAKENHI Grant Number  JP23K18509, Japan. 

\bibliographystyle{splncs04}
\bibliography{myrefs}
\end{document}